\documentclass[sigconf]{acmart}

\AtBeginDocument{%
  }

\begin{document}

\title{Object-Driven One-Shot Fine-tuning of Text-to-Image Diffusion with Prototypical Embedding}

\author{Jianxiang Lu}
\affiliation{%
  \institution{Tencent}
  \country{China}}

\email{jianxianglu2023@gmail.com}

\author{Cong Xie}
\affiliation{%
  \institution{Tencent}
  \country{China}}
\email{xie.cong@outlook.com}

\author{Hui Guo}
\affiliation{%
  \institution{Tencent}
  \country{China}}
\email{emmaguo7033@gmail.com}

\renewcommand\footnotetextcopyrightpermission[1]{}
\settopmatter{printacmref=true} 

\begin{abstract}

  As large-scale text-to-image generation models have made remarkable progress in the field of text-to-image generation, many fine-tuning methods have been proposed. However, these models often struggle with novel objects, especially with one-shot scenarios. Our proposed method aims to address the challenges of generalizability and fidelity in an object-driven way, using only a single input image and the object-specific regions  of interest. To improve generalizability and mitigate overfitting, in our paradigm, a prototypical  embedding  is initialized based on the object's appearance and its class, before fine-tuning the diffusion model. And during fine-tuning, we propose a class-characterizing regularization to preserve prior knowledge of object classes. To further improve fidelity,  we introduce  object-specific loss, which can also use to implant multiple objects. Overall, our proposed object-driven method for implanting new objects can integrate seamlessly with existing concepts as well as with high fidelity and generalization. Our method outperforms several existing works. The code will be released.
\end{abstract}




\begin{teaserfigure}
  \includegraphics[width=\textwidth]{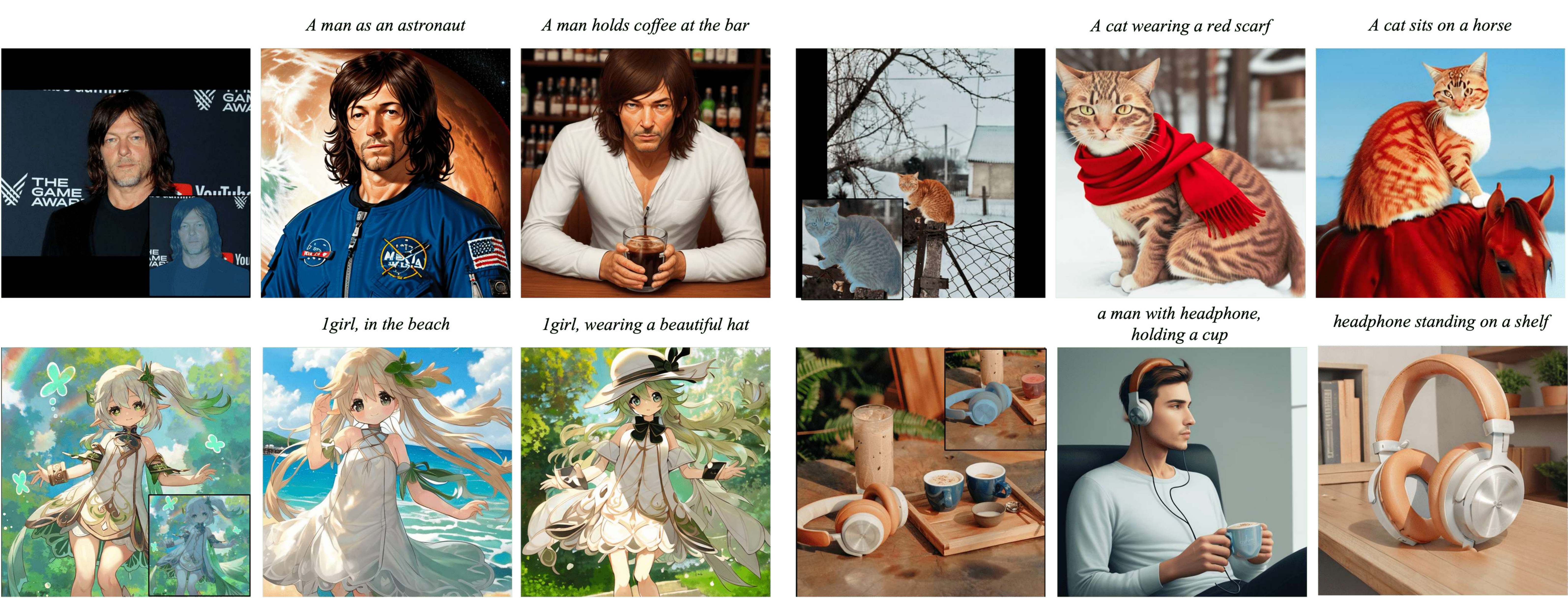}
  \caption{Given an image and its corresponding mask of an object, our method generates diverse images of the same object in different contexts. Moreover, our method allows for the implantation of multiple objects, as demonstrated in the lower right corner of the figure.}
  \label{fig:display}
\end{teaserfigure}


\maketitle

\section{Introduction}

Recently, text-to-image generation using deep learning models have significant development  \cite{ramesh2022hierarchical, rombach2022high, saharia2022photorealistic, yu2022scaling, sauer2023stylegan, chen2022re}. With intricate designs and vast quantities of training data, these models are capable of generating images of unparalleled quality from unrestricted natural language inputs, provided by the user. These systems can create diverse objects, styles, and scenes, as well as generating different object compositions, resulting in aesthetically pleasing outputs.

Apart from general image generation, users are also concerned with whether it is possible to generate images of a specific object in various text descriptions while providing limited samples, such as a single image. This is a one-shot generation task that requires accurate understanding of the specific object and effective model fine-tuning while ensuring the object similarity in generated images and the generation controllability of the resultant model. However, this task poses significant challenges due to limited training samples and a general lack of specific objects in the model training set.


In order to improve the effectiveness of object fine-tuning, current methods \cite{gal2022image, ruiz2022dreambooth, brooks2022instructpix2pix, kumari2022multi, mokady2022null} employ a paradigm that involves optimizing a pre-trained synthesis model, or a text prompt \cite{gal2022image, mokady2022null} representing the object using multiple training examples. The optimized prompt is then integrated with user prompts to create images with diverse styles and content. However, these methods usually rely on the availability of multiple images of the same object, which exacerbates the problem of object accuracy and model overfitting in one-shot scenarios.



Focusing on the task of accurate inplantation of a user specified object into a generative model using only one sample, while maintaining the model's generalization, we propose a novel fine-tuning framework. Unlike existing works \cite{ruiz2022dreambooth, kumari2022multi, mokady2022null, wen2023hard} that start from random initialization and iterative optimize a unique prompt representation of a target object, we propose a novel fine-tuning method. Based on stable diffusion, a prototypical embedding for the object is firstly initialized and then trained in the generation model with a class-characteristic regularization to balance object identity and generalization variations. And an object-specific loss function supervised by the object in the given image is proposed, to achieve both fidelity and generalization in later image synthesis.


While seemingly simple, one-shot generation fine-tuning remains a non-trivial task that poses several challenges, including: 1) adapting text-to-image models with conditioned object embeddings, 2) designing a training process to improve object fidelity using only one image, and 3) enhancing editing capabilities to generate different object compositions using diverse text prompts. Given a target image and its class, the text embedding of the specific object is initialized by finding the best prototypical embedding between multi-modal representations of the object's class and its characteristics. For the fine-tuning quality, we insert additional attention modules\cite{hu2021lora} to the text-to-image network, and an object-specific mask loss is used to preserve fidelity in downstream image synthesis. However, directly fine-tuning the network with the object embedding results in model overfitting with deteriorated editing capability, i.e. high Kernel Inception Distance  (KID) \cite{binkowski2018demystifying} in generation results, where the model generates solely with the user specific object but not the text instruction. We therefore propose a class-characterizing regularization to the object embedding for the generalization of the model.  Our method essentially increases object fidelity and model generalization, leading to performance gain in terms of output quality, appearance diversity, and object fidelity. 

\begin{figure}[ht]
  \centering
  \includegraphics[width=\linewidth]{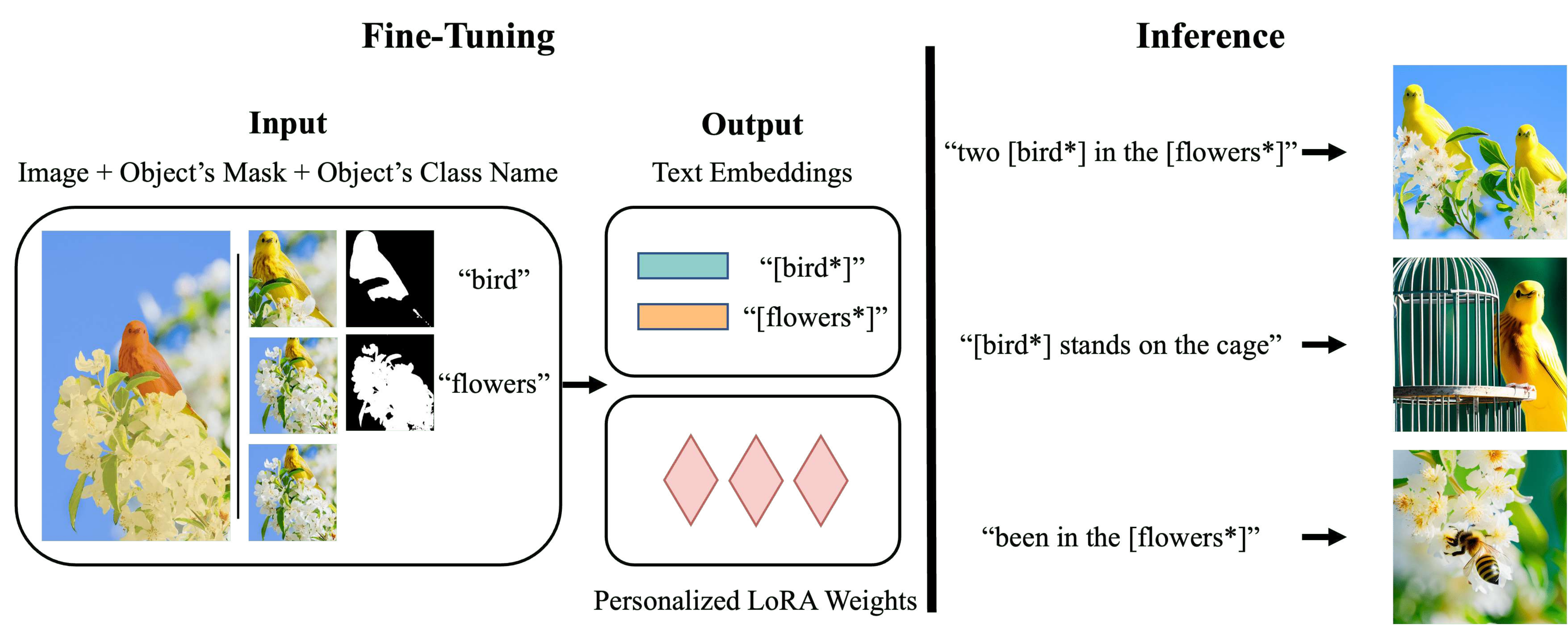}
  \caption{Methodology overview. Our method takes an input image along with its corresponding masks and relevant class names as input, generating object-specific text embeddings and personalized LoRA weights. During inference, the text embedding and LoRA weight is combined with other features to generate a wide range of variations for the object.}
  \label{fig:method-overview}
\end{figure}

Generating personalized content based on one-shot samples requires an efficient algorithm to improve its applicability. In this work, we take a step forward and demonstrate how the fine-tuning process can help the model maintain object identity as long as generalization editability. Our method outperforms several existing methods, which generates high-fidelity photorealistic images and supports content-rich synthesis in one-shot generation. What's more, our method can also used in multi-object fine-tuning, as shown in Figure \ref{fig:method-overview}, which achieve better composing results of multi user specific objects.

\section{Related Work}

\subsection{Text-to-Image Synthesis}
In recent years, there has been significant progress in the development of text-to-image generative models that synthesize images based on unconditioned text prompts. Such models include Generative Adversarial Networks (GANs) \cite{brock2018large, goodfellow2020generative,karras2021alias, sauer2023stylegan}, Vector Quantization (VQs) approaches \cite{razavi2019generating, esser2021taming, yu2022scaling, chang2023muse, ding2022cogview2}, and diffusion models \cite{sohl2015deep, ho2020denoising, song2020denoising, dhariwal2021diffusion, rombach2022high, saharia2022photorealistic,sheynin2022knn, nichol2021glide}. With the rise of large-scale text-image datasets, these models have been scaled up to produce high-quality and semantically rich image synthesis results. Examples of such large text-to-image models include StyleGAN-T \cite{sauer2023stylegan}, CogView2 \cite{ding2022cogview2}, Imagen \cite{saharia2022photorealistic}, DALL-E2 \cite{ramesh2022hierarchical}, and Stable Diffusion \cite{rombach2022high}. The introduction of text conditioning to the StyleGAN architecture in StyleGAN-T has showcased the effectiveness of GANs in text-to-image synthesis. In contrast, VQs utilize an autoencoder to learn a quantization codebook and predict text-to-image tokens through transformers for generating the final image. Likewise, diffusion models leverage a UNet module to iteratively denoise a noise image in a latent space, conditioned on the text prompt injected to the model via transformer layers, to synthesize images.



While text-to-image models trained on a specific dataset can generate high-quality images, they often struggle to generate novel instances or user-specified objects with identity consistency \cite{kumari2022multi}. This can be attributed to their training set limitation in representing new objects. Our work relates to the Stable Diffusion model and aims to enhance its ability to generate high-fidelity images of user-specified objects, thereby expanding the model's applicability to content creation tasks that require personalized text-to-image synthesis.

\subsection{Personalized image Synthesis}
Fine-tuning method with a pre-trained synthesis model for personalized text-to-image generation purposs mainly includes the following four methods: parts of the network parameter tuning \cite{ruiz2022dreambooth, kumari2022multi}, image augmentation based method \cite{esser2021taming, brooks2022instructpix2pix}, prompt tuning or text encoder adapting \cite{gal2022image, mokady2022null} and injecting additional module which adapts to the new objects, i.e., Low-Rank Adaptation (LoRA)  \cite{hu2021lora}. Adapting the whole model or part of the parameters \cite{ruiz2022dreambooth} using a one-shot image can easily lead to overfitting, as it modifies well-trained parameters and can cause catastrophic forgetting \cite{french1999catastrophic, kirkpatrick2017overcoming, li2022overcoming, li2017learning, ramasesh2022effect}. While augmentation \cite{esser2021taming} or prompt tuning \cite{gal2022image} based methods provide a wide variety of samples for regularization or text encoder fine-tuning in training process, they not only struggle to preserve object identity but also require additional training information. Methods by adding an extra module can ease the overfitting problem while sharing the same low fidelity problem. 
In this paper, a small number of parameters, which relates to text encoder and transformer cross attention module, is adapted for the new object and the model is fine-tuned with object-driven method to preserve fidelity and generalization ability of composing new object with the existing ones. 

\begin{figure*}[ht]
  \centering
  \includegraphics[width=\textwidth]{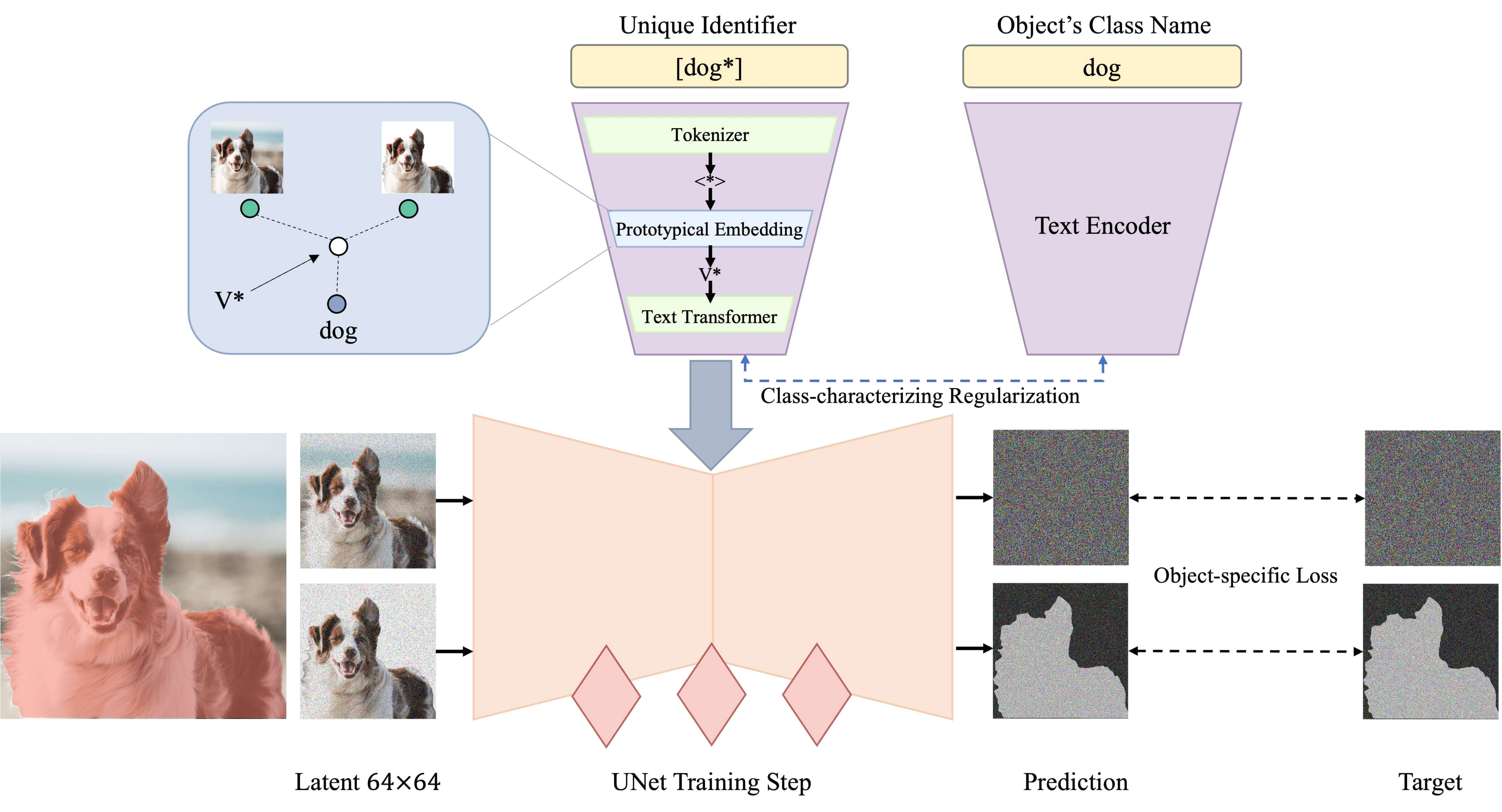}
  \caption{Fine-tuning details. Given one image with single or multiple objects, our method fine-tunes a text-to-image diffusion model. Taking single object as an example, our method utilizes prototypical embedding for initialization and employs class-characterizing regularization to enhance generation diversity, along with a class-specific loss function to ensure fidelity of the synthesized images.}
  \label{fig:method-finetune}
\end{figure*}

\subsection{One-shot Synthesis} 

Fine-tuning a diffusion model with only one sample is an extremely challenging task. While existing fine-tuning methods \cite{ruiz2022dreambooth} or prompt-tuning approaches \cite{gal2022image} suffer from overfitting issues in one-shot scenarios, additional training strategies need to be developed. To maintain generation controllability, Dong et al. \cite{dong2022dreamartist} use positive and negative prompts to contrastively fine-tune the representation of a pseudo prompt, which had generation failures in multi-object compositions. Another method \cite{esser2021taming} considers the image background of the one-shot sample while implanting the object and uses a background-masked object embedding for fine-tuning. It trains the entire network and the text encoder with side information such as original training samples for regularization, which faces the problem of generation defects of the specific object.

In contrast to the aforementioned paradigms, our method only requires one image and the object's region of interest, and focusing on the specific fidelity and generalization variations with specific object synthesis with . It differs from existing methods in several aspects. Firstly, we introduce an object-driven prototypical embedding initialization for the new object, which alleviates the difficulty of representing the object with only one image and improve the efficiency of object inplantation. Secondly, we introduce an object-driven specific loss for precise object appearance learning, where \cite{kirillov2023segment} is used for the object mask. Thirdly, a LoRA \cite{hu2021lora} module where the main denoising UNet is maintained and a class-characteristic regularization to protect class prior information for semantic generalization with other objects and preventing catastrophic forgetting. Moreover, our method is capable of multiple object fine-tuning, which is challenging in existing methods \cite{ruiz2022dreambooth,kumari2022multi}.

\section{Method}

\subsection{Overview}


Our proposed method  focuses on object-driven fine-tuning of single or multiple objects specified by the user in one image, as shown in Figure \ref{fig:method-finetune}. To overcome the limitations of existing fine-tuning methods, we use prototypical embedding as the initialization embedding and propose a regularized loss function to increase  the diversity of the generated images and effectively preserve the prior knowledge of the pre-trained model. In addition, we introduce an object-specific mask loss function to synthesize high-fidelity images, which can also be used for multi-object implantation. In this section, we explain the proposed method in detail.

\begin{figure*}[ht]
  \centering
  \includegraphics[width=\linewidth]{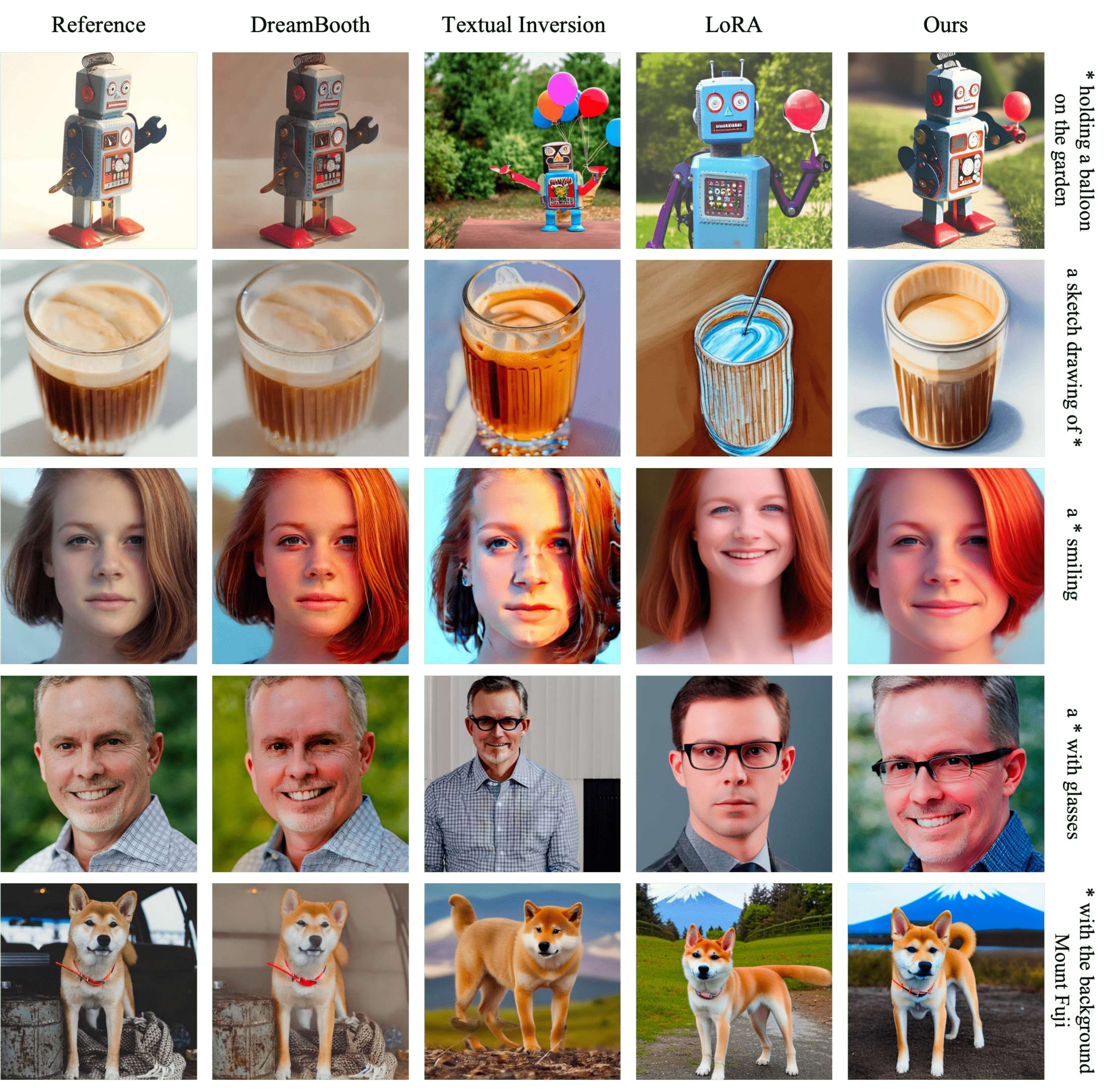}
  \caption{Qualitative comparison. For one-shot tasks, existing methods face challenges in achieving both fidelity and generalizability with the given text. Our method generates images that better match the reference image and are consistent with the text semantics under multiple cue words. Note that the * symbol represents a unique identifier.}
  \label{fig:comp}
\end{figure*}

\subsection{Latent Diffusion Model}


In this work, we adopt the stable diffusion model, a well-established text-to-image diffusion model, as our synthetic network. Stable diffusion is a Latent Diffusion Model (LDM) \cite{rombach2022high}, which performs denoising operations in the latent space instead of the image space. In simple terms, first, the RGB image $x$ is encoded into the latent representation $z=\varepsilon(x)$ by the encoder $\varepsilon$ of an Variable Auto Encoder (VAE). In the latent space, the denoising U-Net of LDM employs the cross-attention mechanism for text-conditional image generation. The training loss of the conditional LDM is formulated as:
\begin{equation}
    L_{L D M}=\mathbb{E}_{\mathcal{E}(x), y, \epsilon \sim \mathcal{N}(0,1), t}\left[\left\|\epsilon-\epsilon_\theta\left(z_t, t, c\right)\right\|_2^2\right]
\end{equation}
where $c$ is the text embedding, $\epsilon$ denotes standard Gaussian, and $\epsilon_\theta$ is the model prediction. $z_t$ is the noisy version of input $z$ in timestep $t$.


\subsection{Prototypical  Embedding}
\label{sec:pe}


When fine-tuning a diffusion model, the text embedding of the object is usually trained. However, when the training data is only one image, it sometimes causes overfitting, leading the network to generate output based only on the text embeddings of the objects, while other textual conditions are ignored. In practice, proper initialization of text embeddings can enable faster fitting of the network and alleviate overfitting, such as Textual Inversion (TI)  \cite{gal2022image} initializing text embeddings based on the class of objects. In this work, in order to achieve more efficient initialization, we find prototype embeddings based on the embedding of the input image and the text embedding of the class name (e.g., dog). Prior to commencing the training of the diffusion model, we compute prototypical embedding via:
\begin{equation}
L_{PE}=1-\frac{\mathcal{T} (c_p) \theta_m(\mathcal{I}(x), \mathcal{I}(x_{m}),  \mathcal{T}(c_c))}{\left \|\mathcal{T} (c_p) \right \| \left \| \theta_m(\mathcal{I}(x), \mathcal{I}(x_{m}),  \mathcal{T}(c_c))  \right \| }
\end{equation}

where $x$ is the training image, the image encoder $\mathcal{I}$ and the text encoder $\mathcal{T}$ of CLIP \cite{radford2021learning} are used to obtain the whole image embedding $\mathcal{I}(x)$ , the object mask image embedding  $\mathcal{I}(x_m)$,  $\mathcal{T} (c_c)$ is the class name text embedding of the object and $\theta_m$ is the way of embedding fusion e.g. averaging. We aim to obtain a prototype text embedding  $\mathcal{T}(c_p)$ similar to the target image embedding and the class text embedding as initialization by this loss function.

\subsection{Class-characterizing Regularization}
Additionally, in order to preserve the synthesis ability of the class of objects in the pre-trained model, we adjust the text embedding using class-characterizing regularization during the training process. Class-characterizing loss is formulated as:

\begin{equation}
    L_{CL}=
    \begin{cases}
    1-\alpha_{cl}  \frac{\mathcal{T} (c_p) \mathcal{T}(c_c)}{\left \|\mathcal{T} (c_p) \right \| \left \|  \mathcal{T}(c_c)  \right \| } &  \text{if } p < p_{cl} \\
    0 &  \text{otherwise}
    
    \end{cases}
\end{equation}
where $\mathcal{T} (c_c)$ is the class name text embedding of the object, $\alpha_{cl}$ represents the weight of the cosine loss, $p \sim Uni(0, 1)$, and $p_{cl}$ is an adjustable threshold.
In this context, it is necessary to predetermine the class name of each object. Further experiments indicate that the introduction of this loss function leads to improved generalizability in synthesis.

\subsection{Object-specific Loss}
\label{sec:osloss}
Our task is to implant the selected objects into the output domain of the model and bind them with a unique identifier. Noted that the selected objects are often parts of the training image rather than the whole image, for this reason we propose object-specific loss for implantation of selected objects with improved fidelity. First, we use an image segmentation algorithm such as SAM\cite{kirillov2023segment} to obtain the mask images $m$ of the objects. The mask images are introduced into the latent space and the training process. For single-object implantation is trained as follow:
\begin{equation}
    L_{SP}=\left\|\tilde{\epsilon}-\epsilon_\theta\left(\tilde{z_t} , t, c_m\right)\right\|^2_2 + \left\|\epsilon-\epsilon_\theta\left(z_t, t, c\right)\right\|^2_2
\end{equation}
where $c_m$ is the text condition of this object with mask, object target noise $\tilde{\epsilon}=\epsilon \otimes m + \epsilon_{\theta} \otimes (1 - m)$, and mask latent representation $\tilde{z}=z \otimes m$. Our goal is to focus on the mask region when performing the loss calculation. Further, for multiple object implantation we make object-specific loss function combinations, assuming that there are a set of $r$ objects to be implanted and a subset $S$ of $k$ distinct objects are taken at a time, the number of $k$-combinations is $C_n^k$. So, in one step of training, the overall object-specific loss is: 
\begin{equation}
L_{SP}= \sum_{i\in S}  \left\|\tilde{\epsilon}_i-\epsilon_\theta\left(\tilde{z}_{t,i}, t, c_{m,i}\right)\right\|^2_2 + \left\|\epsilon-\epsilon_\theta\left(z_t, t, c\right)\right\|^2_2
\end{equation}
Note that the text condition $c_{m,i}$ is different for each mask, and the global text condition $c$ is based on unique identifiers for all objects.


\begin{figure}[htbp]
  \centering
  \includegraphics[width=\linewidth]{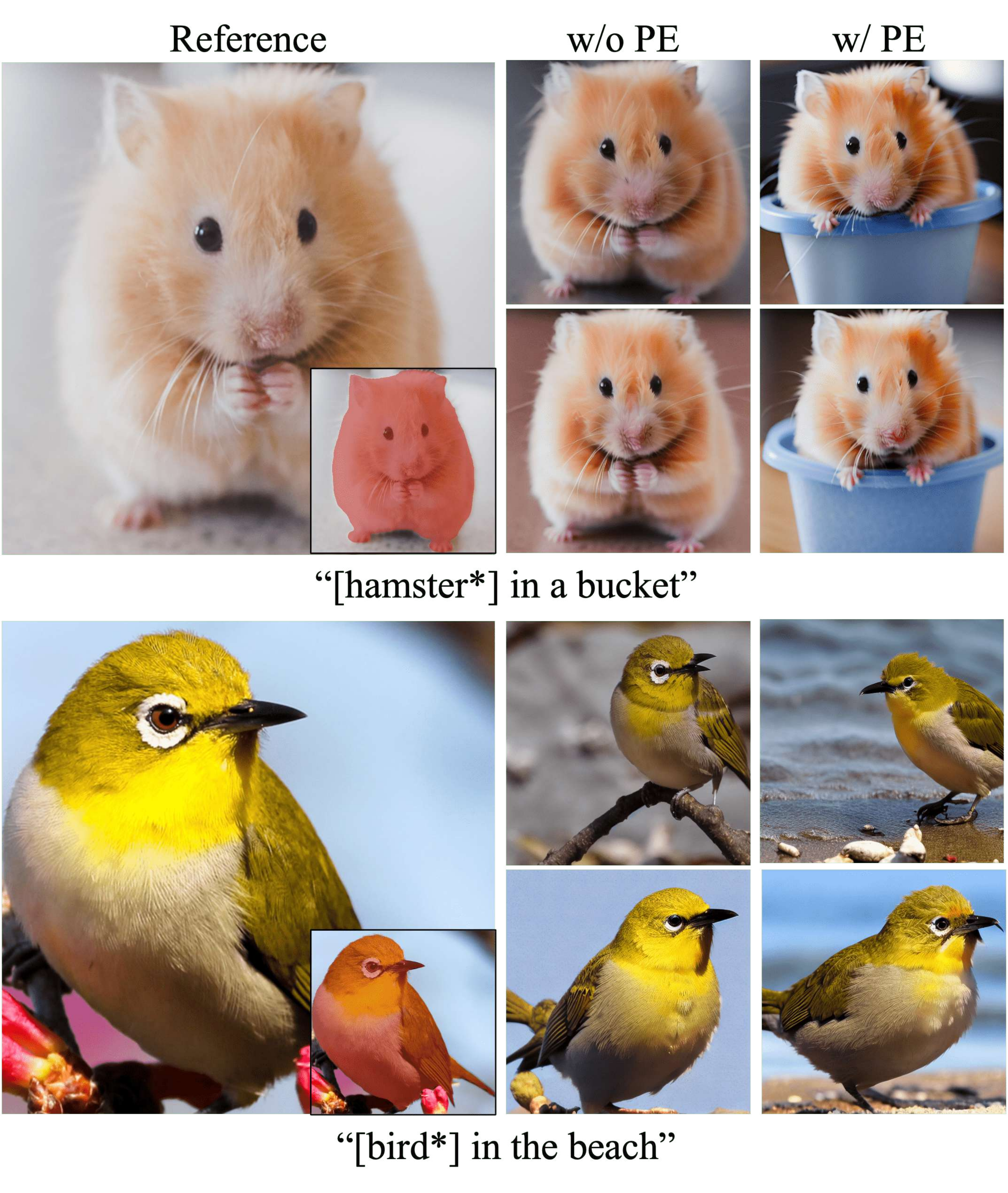}
  \caption{Prototypical embedding initialization. Our proposed method, utilizing prototypical embedding as the initialization, ensures the generation of images that are more contextually relevant.}
  \label{fig:as-pe}
\end{figure}

\section{Experiments}
\subsection{Settings}
We use the pre-trained stable diffusion model as the text-to-image network. The CLIP image and text encoders are used to compute prototypical embeddings and take LoRA as the fine-tuned model. We use the images on unsplash  as fine-tuning data and the mask images of the objects are obtained using SAM. In the object-driven fine-tuning, the learning rate is $10^{-4}$ and the training is 100 steps with a batchsize of 1, on a v100 GPU. We did not over-tune the superparameter, $\alpha_{cl}$, $p_{cl}$ are set to 1, and $k$ is set to 2 for object-specific loss combinations.
We use Dreambooth, TI and LoRA as three state-of-the-art methods for comparison, and use their publicly released implementation code. In addition, their learning rates and training steps are consistent in training. 

\subsection{Comparison}
As shown in Figure \ref{fig:comp}, we compare our results with existing fine-tuning methods. Because of the different fine-tuning strategies, the existing methods may overfit, or not be fidelity, when only one image is given as input. For action generation, the results of TI and LoRA align the actions in the prompt, but the generated objects differ significantly from the reference image. And DreamBooth overfits, resulting in the inability to generate actions. In contrast, our method is able to generate actions with higher fidelity. For human faces, all methods except ours cannot generate expressions and preserve identities at the same time. The same observation is made for animals and style transformation.

\begin{figure}[htbp]
  \centering
  \includegraphics[width=\linewidth]{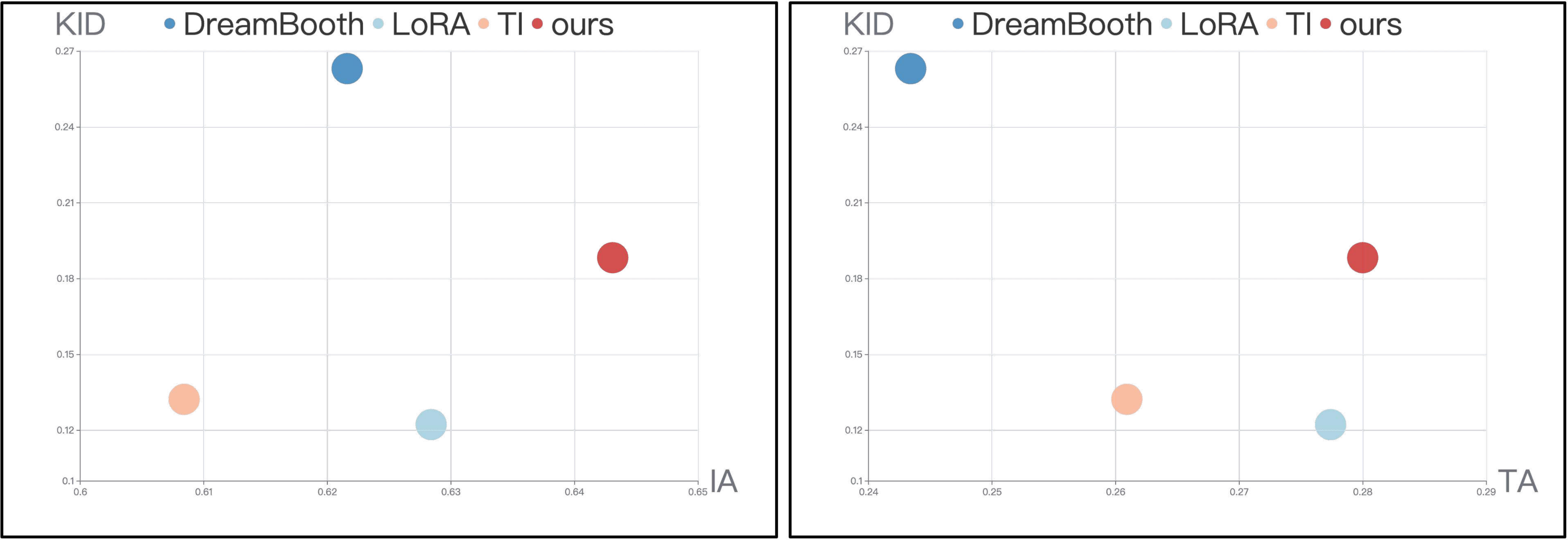}
  \caption{Quantitative assessment. We visualize the metrics for each method, the point towards the lower right, the better performance of the method.}
  \label{fig:comp-points}
\end{figure}

Further, we compare other methods using quantitative metrics.
We quantitatively evaluated 5 categories and 750 images, and the results are tabulated in Table \ref{tab:comp}. we used Text Alignment (TA) \cite{hessel2021clipscore}, Image Alignment (IA) \cite{gal2022image} and kernel inception distance as metrics, where TA is the generalizability of the alignment ability to characterize the method on the new prompt, and IA characterizes the generalizability to the image similarity. Thus, it is a trade-off, as shown in Figure \ref{fig:comp-points}. It can be observed that our proposed method performs well in this trade-off, with both fidelity and generalization.

Overall, our proposed method has both fidelity and generalizability, which indicates that it effectively mitigates the overfitting and learns the characterization of the object.

\begin{table}
  \caption{Quantitative comparison. We use three metrics to evaluate the generalization and fidelity of the fine-tuning method.}
  \begin{tabular}{ccccc}
    \toprule
    Methods & TI & DreamBooth & LoRA & Ours\\
    \midrule
    IA $\uparrow$  & 0.6084 & 0.6216 & 0.6284 & \textbf{0.6431}\\
    TA $\uparrow$ & 0.2609 & 0.2434 & 0.2774 & \textbf{0.2800}\\
    KID $\downarrow$   & 0.1322 & 0.2630 & \textbf{0.1222} & 0.1882\\
  \bottomrule
\end{tabular}
\label{tab:comp}
\end{table}

\subsection{Ablation Study}

\begin{figure}[htbp]
  \centering
  \includegraphics[width=\linewidth]{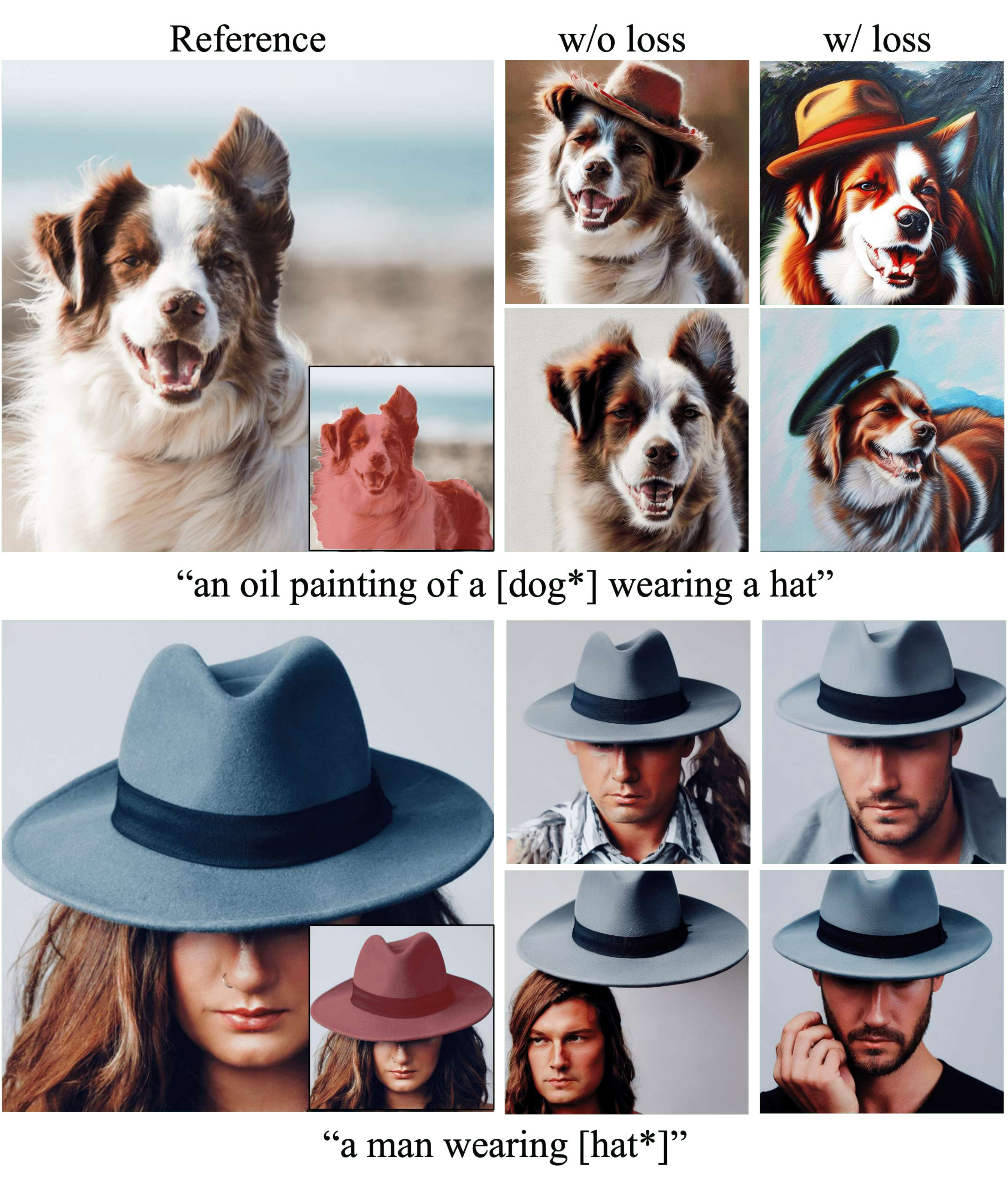}
  \caption{Class-characterizing regularization. Class-characterizing regularization preserves the prior representation of object classes during fine-tuning, resulting in a more natural and diverse synthesis of objects.}
  \label{fig:as-ccr}
\end{figure}

\subsubsection{Prototypical Embedding Initialization} 
As indicated in Section \ref{sec:pe}, we propose prototypical embedding to mitigate the problem of overfitting for  fine-tuning, and to demonstrate its importance we compare the results of random initialization embedding. Note that in all comparisons the initialization of the text embedding is based on four token vectors. As shown in Figure \ref{fig:as-pe}, without the prototypical embedding, the synthesized image has only the reference object and ignores other information in the prompt. On the contrary, when prototypical embedding is adopted, the model is able to generate elements other than objects (e.g., buckets). It demonstrates that prototypical embedding is effective in overcoming the limitations of overfitting and improving the diversity of image generation.

\begin{figure}[htbp]
  \centering
  \includegraphics[width=\linewidth]{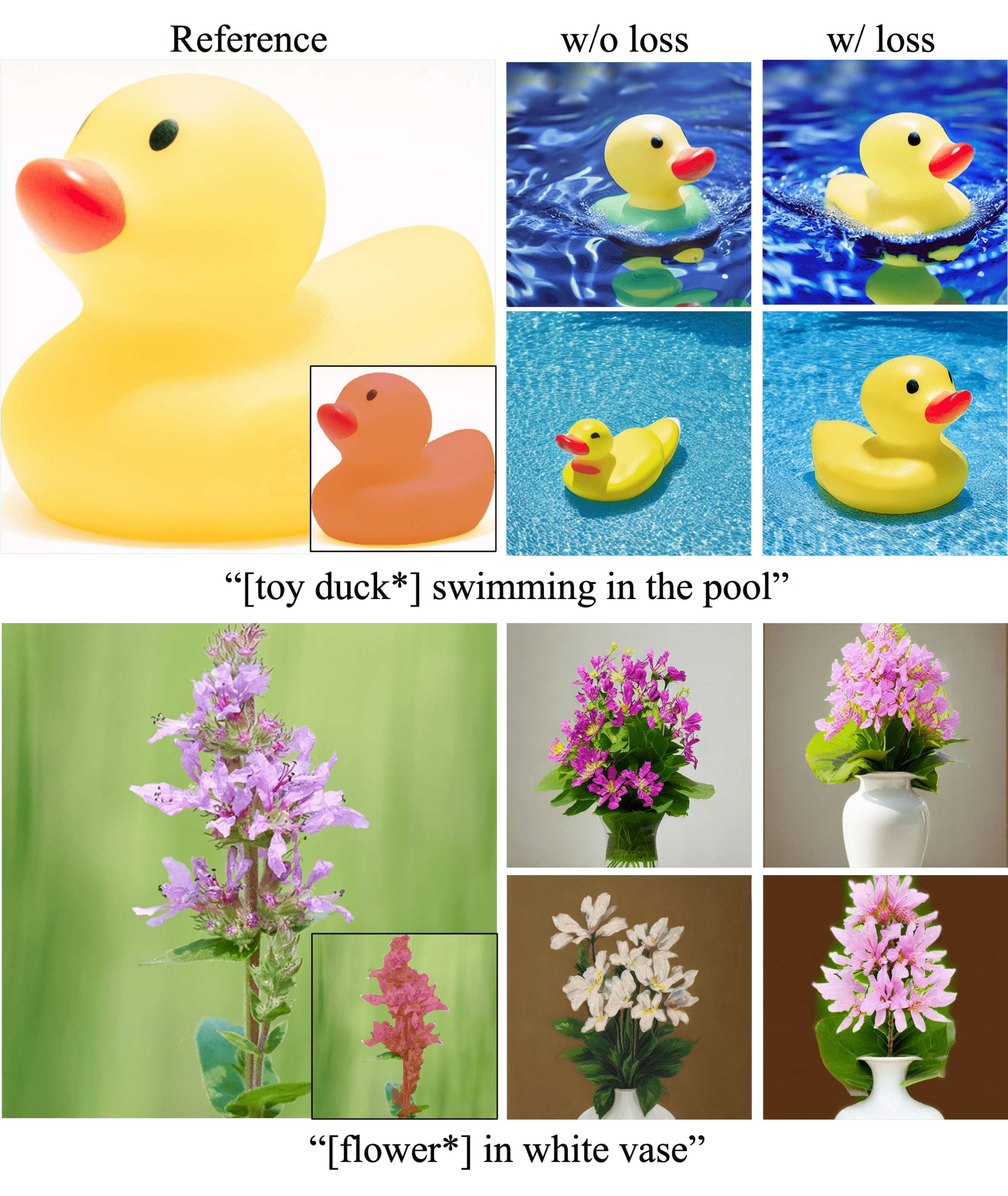}
  \caption{Object-specific loss. Our object-specific loss function enhances attention to the specific object region, thereby effectively preserving the fidelity of the generated objects.}
  \label{fig:as-osloss}
\end{figure}

\subsubsection{Class-characterizing Regularization} 
Since prototypical embedding is only used as a way to initialize without involving the training process, the prior  representations of  object classes are sometimes lost during fine-tuning.  As shown in Figure \ref{fig:as-ccr}, when without class-characterizing regularization, the generated hats are not well integrated with the person, and customized styles cannot be generated. We observe that with class-characterizing regularization, the prior knowledge of the object's class (e.g., hat) is preserved in fine-tuning, and the generated images possess greater diversity and naturalness.

\subsubsection{Object-specific Loss} 
In object-driven training, we adopt an object-specific loss function that focuses on the object region to enhance object fidelity. To demonstrate the effectiveness of the loss, we compare the results without the loss function. In Figure \ref{fig:as-osloss}, we observe that our results trained with the object-specific loss function have greater fidelity in the color, shape, and details of the objects. For example, the color of the object "flower" does not match the reference image without this loss function, and the object "toy duck" exhibits alterations in its shape and details.

\subsubsection{Multi-object Implantation}

\begin{figure}[htbp]
  \centering
  \includegraphics[width=\linewidth]{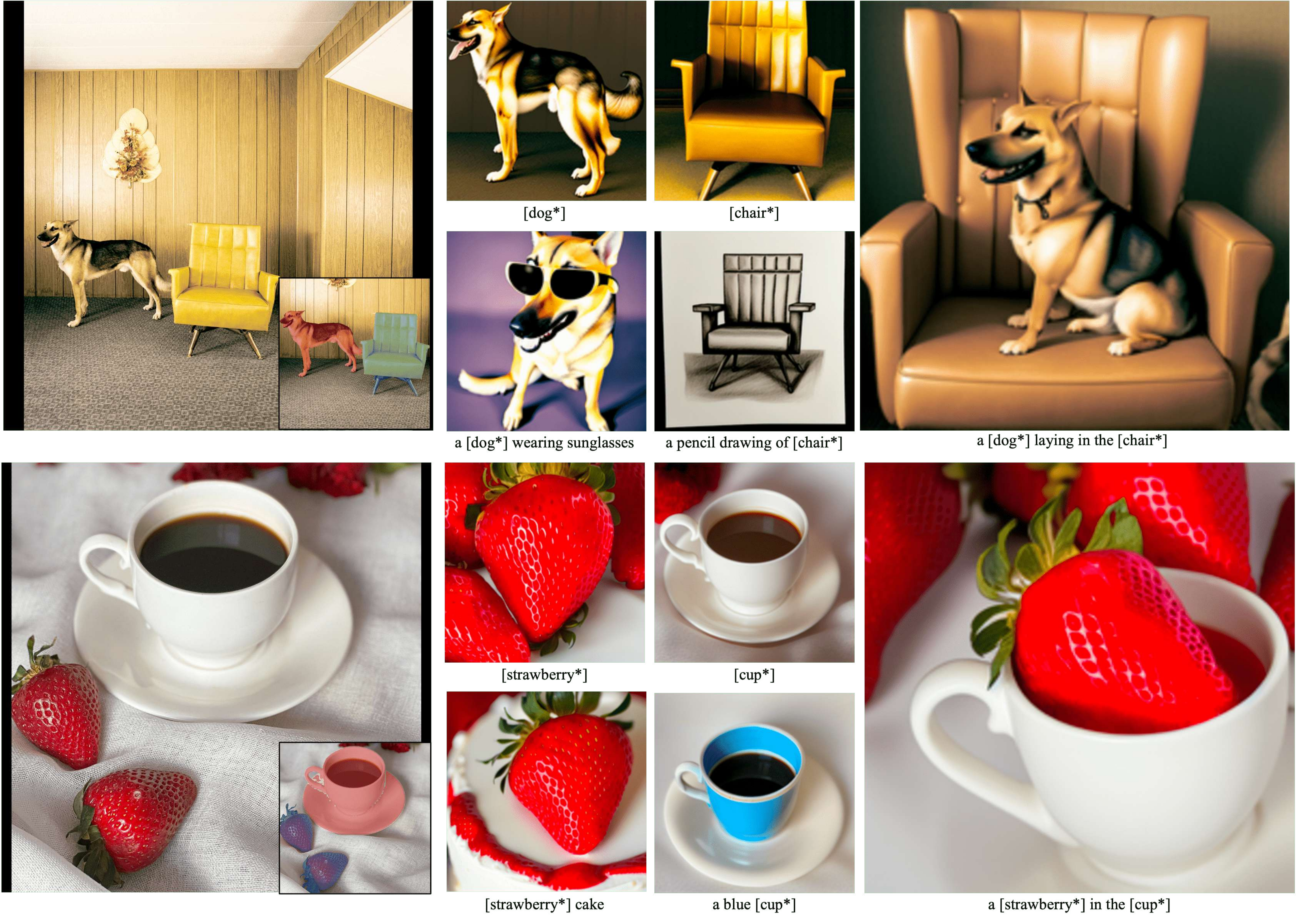}
  \caption{Multi-object implantation. After implanting multiple objects, our method can generate multi objects simultaneously. }
  \label{fig:multi-classes}
\end{figure}

As discussed above, the proposed one-shot method supports multiple objects implantation, which can derive many interesting applications. We first obtain the mask images of the objects separately using the mature image segmentation algorithm and then introduce the object-specific loss combinations mentioned in Section \ref{sec:osloss} for training. As shown in Figure \ref{fig:multi-classes}, the implanted objects do not affect each other when generating images using their unique identifiers and can generate output related to the prompt. In addition, we generate images using multiple object identifiers at the same time (e.g., [strawberry*] in the [cup*]), it can be observed that the output resulting from the combination of concepts is both natural and faithful.

\section{Conclusion}
The main goal of our work is to overcome the challenges of one-shot fine-tuning of text-to-image diffusion, and only one image is required for single or multiple object implantation.  
First we propose the prototypical embedding as an initialization for fine-tuning, and experiments show that it can effectively mitigate overfitting.During training, we introduce class-characterizing regularization to protect the prior knowledge of the pre-trained model and further enhance the synthetic diversity. In addition, the object-specific Loss effective for improving the fidelity of the generated images and can handle multiple objects. Our approach is object-driven and competitive with other existing approaches.  We anticipate that this work will greatly enhance the quality and efficiency of personalized image generation.

At the same time, we also found that our method has some limitations, such as errors in the mask region for objects with complex edges, which sometimes cause degradation in the quality of the generated image edges. In addition, the fidelity of the generated images is slightly degraded when implanting smaller objects. In order to solve the above problems, future work will be devoted to improving the way of acquiring mask images and adding multi-scale perception mechanism for objects.

\bibliographystyle{ACM-Reference-Format}
\bibliography{main}


\begin{thebibliography}{36}


\ifx \showCODEN    \undefined \def \showCODEN     #1{\unskip}     \fi
\ifx \showDOI      \undefined \def \showDOI       #1{#1}\fi
\ifx \showISBNx    \undefined \def \showISBNx     #1{\unskip}     \fi
\ifx \showISBNxiii \undefined \def \showISBNxiii  #1{\unskip}     \fi
\ifx \showISSN     \undefined \def \showISSN      #1{\unskip}     \fi
\ifx \showLCCN     \undefined \def \showLCCN      #1{\unskip}     \fi
\ifx \shownote     \undefined \def \shownote      #1{#1}          \fi
\ifx \showarticletitle \undefined \def \showarticletitle #1{#1}   \fi
\ifx \showURL      \undefined \def \showURL       {\relax}        \fi
\providecommand\bibfield[2]{#2}
\providecommand\bibinfo[2]{#2}
\providecommand\natexlab[1]{#1}
\providecommand\showeprint[2][]{arXiv:#2}

\bibitem[Bi{\'n}kowski et~al\mbox{.}(2018)]%
        {binkowski2018demystifying}
\bibfield{author}{\bibinfo{person}{Miko{\l}aj Bi{\'n}kowski}, \bibinfo{person}{Danica~J Sutherland}, \bibinfo{person}{Michael Arbel}, {and} \bibinfo{person}{Arthur Gretton}.} \bibinfo{year}{2018}\natexlab{}.
\newblock \showarticletitle{Demystifying mmd gans}.
\newblock \bibinfo{journal}{\emph{arXiv preprint arXiv:1801.01401}} (\bibinfo{year}{2018}).
\newblock


\bibitem[Brock et~al\mbox{.}(2018)]%
        {brock2018large}
\bibfield{author}{\bibinfo{person}{Andrew Brock}, \bibinfo{person}{Jeff Donahue}, {and} \bibinfo{person}{Karen Simonyan}.} \bibinfo{year}{2018}\natexlab{}.
\newblock \showarticletitle{Large scale GAN training for high fidelity natural image synthesis}.
\newblock \bibinfo{journal}{\emph{arXiv preprint arXiv:1809.11096}} (\bibinfo{year}{2018}).
\newblock


\bibitem[Brooks et~al\mbox{.}(2022)]%
        {brooks2022instructpix2pix}
\bibfield{author}{\bibinfo{person}{Tim Brooks}, \bibinfo{person}{Aleksander Holynski}, {and} \bibinfo{person}{Alexei~A Efros}.} \bibinfo{year}{2022}\natexlab{}.
\newblock \showarticletitle{Instructpix2pix: Learning to follow image editing instructions}.
\newblock \bibinfo{journal}{\emph{arXiv preprint arXiv:2211.09800}} (\bibinfo{year}{2022}).
\newblock


\bibitem[Chang et~al\mbox{.}(2023)]%
        {chang2023muse}
\bibfield{author}{\bibinfo{person}{Huiwen Chang}, \bibinfo{person}{Han Zhang}, \bibinfo{person}{Jarred Barber}, \bibinfo{person}{AJ Maschinot}, \bibinfo{person}{Jose Lezama}, \bibinfo{person}{Lu Jiang}, \bibinfo{person}{Ming-Hsuan Yang}, \bibinfo{person}{Kevin Murphy}, \bibinfo{person}{William~T Freeman}, \bibinfo{person}{Michael Rubinstein}, {et~al\mbox{.}}} \bibinfo{year}{2023}\natexlab{}.
\newblock \showarticletitle{Muse: Text-To-Image Generation via Masked Generative Transformers}.
\newblock \bibinfo{journal}{\emph{arXiv preprint arXiv:2301.00704}} (\bibinfo{year}{2023}).
\newblock


\bibitem[Chen et~al\mbox{.}(2022)]%
        {chen2022re}
\bibfield{author}{\bibinfo{person}{Wenhu Chen}, \bibinfo{person}{Hexiang Hu}, \bibinfo{person}{Chitwan Saharia}, {and} \bibinfo{person}{William~W Cohen}.} \bibinfo{year}{2022}\natexlab{}.
\newblock \showarticletitle{Re-imagen: Retrieval-augmented text-to-image generator}.
\newblock \bibinfo{journal}{\emph{arXiv preprint arXiv:2209.14491}} (\bibinfo{year}{2022}).
\newblock


\bibitem[Dhariwal and Nichol(2021)]%
        {dhariwal2021diffusion}
\bibfield{author}{\bibinfo{person}{Prafulla Dhariwal} {and} \bibinfo{person}{Alexander Nichol}.} \bibinfo{year}{2021}\natexlab{}.
\newblock \showarticletitle{Diffusion models beat gans on image synthesis}.
\newblock \bibinfo{journal}{\emph{Advances in Neural Information Processing Systems}}  \bibinfo{volume}{34} (\bibinfo{year}{2021}), \bibinfo{pages}{8780--8794}.
\newblock


\bibitem[Ding et~al\mbox{.}(2022)]%
        {ding2022cogview2}
\bibfield{author}{\bibinfo{person}{Ming Ding}, \bibinfo{person}{Wendi Zheng}, \bibinfo{person}{Wenyi Hong}, {and} \bibinfo{person}{Jie Tang}.} \bibinfo{year}{2022}\natexlab{}.
\newblock \showarticletitle{Cogview2: Faster and better text-to-image generation via hierarchical transformers}.
\newblock \bibinfo{journal}{\emph{arXiv preprint arXiv:2204.14217}} (\bibinfo{year}{2022}).
\newblock


\bibitem[Dong et~al\mbox{.}(2022)]%
        {dong2022dreamartist}
\bibfield{author}{\bibinfo{person}{Ziyi Dong}, \bibinfo{person}{Pengxu Wei}, {and} \bibinfo{person}{Liang Lin}.} \bibinfo{year}{2022}\natexlab{}.
\newblock \showarticletitle{Dreamartist: Towards controllable one-shot text-to-image generation via contrastive prompt-tuning}.
\newblock \bibinfo{journal}{\emph{arXiv preprint arXiv:2211.11337}} (\bibinfo{year}{2022}).
\newblock


\bibitem[Esser et~al\mbox{.}(2021)]%
        {esser2021taming}
\bibfield{author}{\bibinfo{person}{Patrick Esser}, \bibinfo{person}{Robin Rombach}, {and} \bibinfo{person}{Bjorn Ommer}.} \bibinfo{year}{2021}\natexlab{}.
\newblock \showarticletitle{Taming transformers for high-resolution image synthesis}. In \bibinfo{booktitle}{\emph{Proceedings of the IEEE/CVF conference on computer vision and pattern recognition}}. \bibinfo{pages}{12873--12883}.
\newblock


\bibitem[French(1999)]%
        {french1999catastrophic}
\bibfield{author}{\bibinfo{person}{Robert~M French}.} \bibinfo{year}{1999}\natexlab{}.
\newblock \showarticletitle{Catastrophic forgetting in connectionist networks}.
\newblock \bibinfo{journal}{\emph{Trends in cognitive sciences}} \bibinfo{volume}{3}, \bibinfo{number}{4} (\bibinfo{year}{1999}), \bibinfo{pages}{128--135}.
\newblock


\bibitem[Gal et~al\mbox{.}(2022)]%
        {gal2022image}
\bibfield{author}{\bibinfo{person}{Rinon Gal}, \bibinfo{person}{Yuval Alaluf}, \bibinfo{person}{Yuval Atzmon}, \bibinfo{person}{Or Patashnik}, \bibinfo{person}{Amit~H Bermano}, \bibinfo{person}{Gal Chechik}, {and} \bibinfo{person}{Daniel Cohen-Or}.} \bibinfo{year}{2022}\natexlab{}.
\newblock \showarticletitle{An image is worth one word: Personalizing text-to-image generation using textual inversion}.
\newblock \bibinfo{journal}{\emph{arXiv preprint arXiv:2208.01618}} (\bibinfo{year}{2022}).
\newblock


\bibitem[Goodfellow et~al\mbox{.}(2020)]%
        {goodfellow2020generative}
\bibfield{author}{\bibinfo{person}{Ian Goodfellow}, \bibinfo{person}{Jean Pouget-Abadie}, \bibinfo{person}{Mehdi Mirza}, \bibinfo{person}{Bing Xu}, \bibinfo{person}{David Warde-Farley}, \bibinfo{person}{Sherjil Ozair}, \bibinfo{person}{Aaron Courville}, {and} \bibinfo{person}{Yoshua Bengio}.} \bibinfo{year}{2020}\natexlab{}.
\newblock \showarticletitle{Generative adversarial networks}.
\newblock \bibinfo{journal}{\emph{Commun. ACM}} \bibinfo{volume}{63}, \bibinfo{number}{11} (\bibinfo{year}{2020}), \bibinfo{pages}{139--144}.
\newblock


\bibitem[Hessel et~al\mbox{.}(2021)]%
        {hessel2021clipscore}
\bibfield{author}{\bibinfo{person}{Jack Hessel}, \bibinfo{person}{Ari Holtzman}, \bibinfo{person}{Maxwell Forbes}, \bibinfo{person}{Ronan~Le Bras}, {and} \bibinfo{person}{Yejin Choi}.} \bibinfo{year}{2021}\natexlab{}.
\newblock \showarticletitle{Clipscore: A reference-free evaluation metric for image captioning}.
\newblock \bibinfo{journal}{\emph{arXiv preprint arXiv:2104.08718}} (\bibinfo{year}{2021}).
\newblock


\bibitem[Ho et~al\mbox{.}(2020)]%
        {ho2020denoising}
\bibfield{author}{\bibinfo{person}{Jonathan Ho}, \bibinfo{person}{Ajay Jain}, {and} \bibinfo{person}{Pieter Abbeel}.} \bibinfo{year}{2020}\natexlab{}.
\newblock \showarticletitle{Denoising diffusion probabilistic models}.
\newblock \bibinfo{journal}{\emph{Advances in Neural Information Processing Systems}}  \bibinfo{volume}{33} (\bibinfo{year}{2020}), \bibinfo{pages}{6840--6851}.
\newblock


\bibitem[Hu et~al\mbox{.}(2021)]%
        {hu2021lora}
\bibfield{author}{\bibinfo{person}{Edward~J Hu}, \bibinfo{person}{Yelong Shen}, \bibinfo{person}{Phillip Wallis}, \bibinfo{person}{Zeyuan Allen-Zhu}, \bibinfo{person}{Yuanzhi Li}, \bibinfo{person}{Shean Wang}, \bibinfo{person}{Lu Wang}, {and} \bibinfo{person}{Weizhu Chen}.} \bibinfo{year}{2021}\natexlab{}.
\newblock \showarticletitle{Lora: Low-rank adaptation of large language models}.
\newblock \bibinfo{journal}{\emph{arXiv preprint arXiv:2106.09685}} (\bibinfo{year}{2021}).
\newblock


\bibitem[Karras et~al\mbox{.}(2021)]%
        {karras2021alias}
\bibfield{author}{\bibinfo{person}{Tero Karras}, \bibinfo{person}{Miika Aittala}, \bibinfo{person}{Samuli Laine}, \bibinfo{person}{Erik H{\"a}rk{\"o}nen}, \bibinfo{person}{Janne Hellsten}, \bibinfo{person}{Jaakko Lehtinen}, {and} \bibinfo{person}{Timo Aila}.} \bibinfo{year}{2021}\natexlab{}.
\newblock \showarticletitle{Alias-free generative adversarial networks}.
\newblock \bibinfo{journal}{\emph{Advances in Neural Information Processing Systems}}  \bibinfo{volume}{34} (\bibinfo{year}{2021}), \bibinfo{pages}{852--863}.
\newblock


\bibitem[Kirillov et~al\mbox{.}(2023)]%
        {kirillov2023segment}
\bibfield{author}{\bibinfo{person}{Alexander Kirillov}, \bibinfo{person}{Eric Mintun}, \bibinfo{person}{Nikhila Ravi}, \bibinfo{person}{Hanzi Mao}, \bibinfo{person}{Chloe Rolland}, \bibinfo{person}{Laura Gustafson}, \bibinfo{person}{Tete Xiao}, \bibinfo{person}{Spencer Whitehead}, \bibinfo{person}{Alexander~C Berg}, \bibinfo{person}{Wan-Yen Lo}, {et~al\mbox{.}}} \bibinfo{year}{2023}\natexlab{}.
\newblock \showarticletitle{Segment anything}.
\newblock \bibinfo{journal}{\emph{arXiv preprint arXiv:2304.02643}} (\bibinfo{year}{2023}).
\newblock


\bibitem[Kirkpatrick et~al\mbox{.}(2017)]%
        {kirkpatrick2017overcoming}
\bibfield{author}{\bibinfo{person}{James Kirkpatrick}, \bibinfo{person}{Razvan Pascanu}, \bibinfo{person}{Neil Rabinowitz}, \bibinfo{person}{Joel Veness}, \bibinfo{person}{Guillaume Desjardins}, \bibinfo{person}{Andrei~A Rusu}, \bibinfo{person}{Kieran Milan}, \bibinfo{person}{John Quan}, \bibinfo{person}{Tiago Ramalho}, \bibinfo{person}{Agnieszka Grabska-Barwinska}, {et~al\mbox{.}}} \bibinfo{year}{2017}\natexlab{}.
\newblock \showarticletitle{Overcoming catastrophic forgetting in neural networks}.
\newblock \bibinfo{journal}{\emph{Proceedings of the national academy of sciences}} \bibinfo{volume}{114}, \bibinfo{number}{13} (\bibinfo{year}{2017}), \bibinfo{pages}{3521--3526}.
\newblock


\bibitem[Kumari et~al\mbox{.}(2022)]%
        {kumari2022multi}
\bibfield{author}{\bibinfo{person}{Nupur Kumari}, \bibinfo{person}{Bingliang Zhang}, \bibinfo{person}{Richard Zhang}, \bibinfo{person}{Eli Shechtman}, {and} \bibinfo{person}{Jun-Yan Zhu}.} \bibinfo{year}{2022}\natexlab{}.
\newblock \showarticletitle{Multi-Concept Customization of Text-to-Image Diffusion}.
\newblock \bibinfo{journal}{\emph{arXiv preprint arXiv:2212.04488}} (\bibinfo{year}{2022}).
\newblock


\bibitem[Li et~al\mbox{.}(2022)]%
        {li2022overcoming}
\bibfield{author}{\bibinfo{person}{Dingcheng Li}, \bibinfo{person}{Zheng Chen}, \bibinfo{person}{Eunah Cho}, \bibinfo{person}{Jie Hao}, \bibinfo{person}{Xiaohu Liu}, \bibinfo{person}{Fan Xing}, \bibinfo{person}{Chenlei Guo}, {and} \bibinfo{person}{Yang Liu}.} \bibinfo{year}{2022}\natexlab{}.
\newblock \showarticletitle{Overcoming catastrophic forgetting during domain adaptation of seq2seq language generation}. In \bibinfo{booktitle}{\emph{Proceedings of the 2022 Conference of the North American Chapter of the Association for Computational Linguistics: Human Language Technologies}}. \bibinfo{pages}{5441--5454}.
\newblock


\bibitem[Li and Hoiem(2017)]%
        {li2017learning}
\bibfield{author}{\bibinfo{person}{Zhizhong Li} {and} \bibinfo{person}{Derek Hoiem}.} \bibinfo{year}{2017}\natexlab{}.
\newblock \showarticletitle{Learning without forgetting}.
\newblock \bibinfo{journal}{\emph{IEEE transactions on pattern analysis and machine intelligence}} \bibinfo{volume}{40}, \bibinfo{number}{12} (\bibinfo{year}{2017}), \bibinfo{pages}{2935--2947}.
\newblock


\bibitem[Mokady et~al\mbox{.}(2022)]%
        {mokady2022null}
\bibfield{author}{\bibinfo{person}{Ron Mokady}, \bibinfo{person}{Amir Hertz}, \bibinfo{person}{Kfir Aberman}, \bibinfo{person}{Yael Pritch}, {and} \bibinfo{person}{Daniel Cohen-Or}.} \bibinfo{year}{2022}\natexlab{}.
\newblock \showarticletitle{Null-text Inversion for Editing Real Images using Guided Diffusion Models}.
\newblock \bibinfo{journal}{\emph{arXiv preprint arXiv:2211.09794}} (\bibinfo{year}{2022}).
\newblock


\bibitem[Nichol et~al\mbox{.}(2021)]%
        {nichol2021glide}
\bibfield{author}{\bibinfo{person}{Alex Nichol}, \bibinfo{person}{Prafulla Dhariwal}, \bibinfo{person}{Aditya Ramesh}, \bibinfo{person}{Pranav Shyam}, \bibinfo{person}{Pamela Mishkin}, \bibinfo{person}{Bob McGrew}, \bibinfo{person}{Ilya Sutskever}, {and} \bibinfo{person}{Mark Chen}.} \bibinfo{year}{2021}\natexlab{}.
\newblock \showarticletitle{Glide: Towards photorealistic image generation and editing with text-guided diffusion models}.
\newblock \bibinfo{journal}{\emph{arXiv preprint arXiv:2112.10741}} (\bibinfo{year}{2021}).
\newblock


\bibitem[Radford et~al\mbox{.}(2021)]%
        {radford2021learning}
\bibfield{author}{\bibinfo{person}{Alec Radford}, \bibinfo{person}{Jong~Wook Kim}, \bibinfo{person}{Chris Hallacy}, \bibinfo{person}{Aditya Ramesh}, \bibinfo{person}{Gabriel Goh}, \bibinfo{person}{Sandhini Agarwal}, \bibinfo{person}{Girish Sastry}, \bibinfo{person}{Amanda Askell}, \bibinfo{person}{Pamela Mishkin}, \bibinfo{person}{Jack Clark}, {et~al\mbox{.}}} \bibinfo{year}{2021}\natexlab{}.
\newblock \showarticletitle{Learning transferable visual models from natural language supervision}. In \bibinfo{booktitle}{\emph{International conference on machine learning}}. PMLR, \bibinfo{pages}{8748--8763}.
\newblock


\bibitem[Ramasesh et~al\mbox{.}(2022)]%
        {ramasesh2022effect}
\bibfield{author}{\bibinfo{person}{Vinay~Venkatesh Ramasesh}, \bibinfo{person}{Aitor Lewkowycz}, {and} \bibinfo{person}{Ethan Dyer}.} \bibinfo{year}{2022}\natexlab{}.
\newblock \showarticletitle{Effect of scale on catastrophic forgetting in neural networks}. In \bibinfo{booktitle}{\emph{International Conference on Learning Representations}}.
\newblock


\bibitem[Ramesh et~al\mbox{.}(2022)]%
        {ramesh2022hierarchical}
\bibfield{author}{\bibinfo{person}{Aditya Ramesh}, \bibinfo{person}{Prafulla Dhariwal}, \bibinfo{person}{Alex Nichol}, \bibinfo{person}{Casey Chu}, {and} \bibinfo{person}{Mark Chen}.} \bibinfo{year}{2022}\natexlab{}.
\newblock \showarticletitle{Hierarchical text-conditional image generation with clip latents}.
\newblock \bibinfo{journal}{\emph{arXiv preprint arXiv:2204.06125}} (\bibinfo{year}{2022}).
\newblock


\bibitem[Razavi et~al\mbox{.}(2019)]%
        {razavi2019generating}
\bibfield{author}{\bibinfo{person}{Ali Razavi}, \bibinfo{person}{Aaron Van~den Oord}, {and} \bibinfo{person}{Oriol Vinyals}.} \bibinfo{year}{2019}\natexlab{}.
\newblock \showarticletitle{Generating diverse high-fidelity images with vq-vae-2}.
\newblock \bibinfo{journal}{\emph{Advances in neural information processing systems}}  \bibinfo{volume}{32} (\bibinfo{year}{2019}).
\newblock


\bibitem[Rombach et~al\mbox{.}(2022)]%
        {rombach2022high}
\bibfield{author}{\bibinfo{person}{Robin Rombach}, \bibinfo{person}{Andreas Blattmann}, \bibinfo{person}{Dominik Lorenz}, \bibinfo{person}{Patrick Esser}, {and} \bibinfo{person}{Bj{\"o}rn Ommer}.} \bibinfo{year}{2022}\natexlab{}.
\newblock \showarticletitle{High-resolution image synthesis with latent diffusion models}. In \bibinfo{booktitle}{\emph{Proceedings of the IEEE/CVF Conference on Computer Vision and Pattern Recognition}}. \bibinfo{pages}{10684--10695}.
\newblock


\bibitem[Ruiz et~al\mbox{.}(2022)]%
        {ruiz2022dreambooth}
\bibfield{author}{\bibinfo{person}{Nataniel Ruiz}, \bibinfo{person}{Yuanzhen Li}, \bibinfo{person}{Varun Jampani}, \bibinfo{person}{Yael Pritch}, \bibinfo{person}{Michael Rubinstein}, {and} \bibinfo{person}{Kfir Aberman}.} \bibinfo{year}{2022}\natexlab{}.
\newblock \showarticletitle{Dreambooth: Fine tuning text-to-image diffusion models for subject-driven generation}.
\newblock \bibinfo{journal}{\emph{arXiv preprint arXiv:2208.12242}} (\bibinfo{year}{2022}).
\newblock


\bibitem[Saharia et~al\mbox{.}(2022)]%
        {saharia2022photorealistic}
\bibfield{author}{\bibinfo{person}{Chitwan Saharia}, \bibinfo{person}{William Chan}, \bibinfo{person}{Saurabh Saxena}, \bibinfo{person}{Lala Li}, \bibinfo{person}{Jay Whang}, \bibinfo{person}{Emily~L Denton}, \bibinfo{person}{Kamyar Ghasemipour}, \bibinfo{person}{Raphael Gontijo~Lopes}, \bibinfo{person}{Burcu Karagol~Ayan}, \bibinfo{person}{Tim Salimans}, {et~al\mbox{.}}} \bibinfo{year}{2022}\natexlab{}.
\newblock \showarticletitle{Photorealistic text-to-image diffusion models with deep language understanding}.
\newblock \bibinfo{journal}{\emph{Advances in Neural Information Processing Systems}}  \bibinfo{volume}{35} (\bibinfo{year}{2022}), \bibinfo{pages}{36479--36494}.
\newblock


\bibitem[Sauer et~al\mbox{.}(2023)]%
        {sauer2023stylegan}
\bibfield{author}{\bibinfo{person}{Axel Sauer}, \bibinfo{person}{Tero Karras}, \bibinfo{person}{Samuli Laine}, \bibinfo{person}{Andreas Geiger}, {and} \bibinfo{person}{Timo Aila}.} \bibinfo{year}{2023}\natexlab{}.
\newblock \showarticletitle{Stylegan-t: Unlocking the power of gans for fast large-scale text-to-image synthesis}.
\newblock \bibinfo{journal}{\emph{arXiv preprint arXiv:2301.09515}} (\bibinfo{year}{2023}).
\newblock


\bibitem[Sheynin et~al\mbox{.}(2022)]%
        {sheynin2022knn}
\bibfield{author}{\bibinfo{person}{Shelly Sheynin}, \bibinfo{person}{Oron Ashual}, \bibinfo{person}{Adam Polyak}, \bibinfo{person}{Uriel Singer}, \bibinfo{person}{Oran Gafni}, \bibinfo{person}{Eliya Nachmani}, {and} \bibinfo{person}{Yaniv Taigman}.} \bibinfo{year}{2022}\natexlab{}.
\newblock \showarticletitle{Knn-diffusion: Image generation via large-scale retrieval}.
\newblock \bibinfo{journal}{\emph{arXiv preprint arXiv:2204.02849}} (\bibinfo{year}{2022}).
\newblock


\bibitem[Sohl-Dickstein et~al\mbox{.}(2015)]%
        {sohl2015deep}
\bibfield{author}{\bibinfo{person}{Jascha Sohl-Dickstein}, \bibinfo{person}{Eric Weiss}, \bibinfo{person}{Niru Maheswaranathan}, {and} \bibinfo{person}{Surya Ganguli}.} \bibinfo{year}{2015}\natexlab{}.
\newblock \showarticletitle{Deep unsupervised learning using nonequilibrium thermodynamics}. In \bibinfo{booktitle}{\emph{International Conference on Machine Learning}}. PMLR, \bibinfo{pages}{2256--2265}.
\newblock


\bibitem[Song et~al\mbox{.}(2020)]%
        {song2020denoising}
\bibfield{author}{\bibinfo{person}{Jiaming Song}, \bibinfo{person}{Chenlin Meng}, {and} \bibinfo{person}{Stefano Ermon}.} \bibinfo{year}{2020}\natexlab{}.
\newblock \showarticletitle{Denoising diffusion implicit models}.
\newblock \bibinfo{journal}{\emph{arXiv preprint arXiv:2010.02502}} (\bibinfo{year}{2020}).
\newblock


\bibitem[Wen et~al\mbox{.}(2023)]%
        {wen2023hard}
\bibfield{author}{\bibinfo{person}{Yuxin Wen}, \bibinfo{person}{Neel Jain}, \bibinfo{person}{John Kirchenbauer}, \bibinfo{person}{Micah Goldblum}, \bibinfo{person}{Jonas Geiping}, {and} \bibinfo{person}{Tom Goldstein}.} \bibinfo{year}{2023}\natexlab{}.
\newblock \showarticletitle{Hard prompts made easy: Gradient-based discrete optimization for prompt tuning and discovery}.
\newblock \bibinfo{journal}{\emph{arXiv preprint arXiv:2302.03668}} (\bibinfo{year}{2023}).
\newblock


\bibitem[Yu et~al\mbox{.}(2022)]%
        {yu2022scaling}
\bibfield{author}{\bibinfo{person}{Jiahui Yu}, \bibinfo{person}{Yuanzhong Xu}, \bibinfo{person}{Jing~Yu Koh}, \bibinfo{person}{Thang Luong}, \bibinfo{person}{Gunjan Baid}, \bibinfo{person}{Zirui Wang}, \bibinfo{person}{Vijay Vasudevan}, \bibinfo{person}{Alexander Ku}, \bibinfo{person}{Yinfei Yang}, \bibinfo{person}{Burcu~Karagol Ayan}, {et~al\mbox{.}}} \bibinfo{year}{2022}\natexlab{}.
\newblock \showarticletitle{Scaling autoregressive models for content-rich text-to-image generation}.
\newblock \bibinfo{journal}{\emph{arXiv preprint arXiv:2206.10789}} (\bibinfo{year}{2022}).
\newblock


\end{thebibliography}

\end{document}